\documentclass[mnsc]{informs3} 
\usepackage{helvet}
\usepackage{courier}
\usepackage{amssymb}
\usepackage{pifont}
\usepackage{color}
\usepackage{tikz}
\usepackage{makeidx}  
\usepackage{multirow}      
\usepackage{subfigure}
\usepackage{epstopdf}
\usepackage{subfig}
\usepackage{amsmath,dsfont}
\usepackage[utf8]{inputenc}
\usepackage{booktabs, caption, makecell}

\usepackage{threeparttable}
\usepackage{amsfonts}
\usepackage{graphicx,epstopdf,caption}
\usepackage{enumitem}
\usepackage{color}
\usepackage{url}
\usepackage{bbding}
\usepackage{wasysym}
\usepackage{colonequals}
\usepackage{psfrag}
\usepackage{mathrsfs}
\usepackage{lscape}
\usepackage{pifont,textcomp,bbm}
\usepackage{algorithm, algpseudocode}
\usepackage{dsfont}
\usepackage{amsmath,calc}
\usepackage{multicol}
\usepackage{booktabs}

\DoubleSpacedXI 

\usepackage{natbib}
 \bibpunct[, ]{(}{)}{,}{a}{}{,}%
 \def\bibfont{\small}%
\begin{document}




\TITLE{Early Predictions for Medical Crowdfunding: A Deep Learning Approach Using Diverse Inputs}

\ARTICLEAUTHORS{%
\AUTHOR{Tong Wang}
\AFF{University of Iowa}
\AUTHOR{Fujie Jin}
\AFF{Indiana University}
\AUTHOR{Yu (Jeffrey) Hu}
\AFF{Georgia Institute of Technoloty}
\AUTHOR{Yuan Cheng}
\AFF{Tsinghua University}
} 

\ABSTRACT{%
Medical crowdfunding is a popular channel for people needing financial help paying medical bills to collect donations from large numbers of people. However, large heterogeneity exists in donations across cases, and fundraisers face significant uncertainty in whether their crowdfunding campaigns can meet fundraising goals. Therefore, it is important to provide early warnings for fundraisers if such a channel will eventually fail. In this study, we aim to develop novel algorithms to provide accurate and timely predictions of fundraising performance, to better inform fundraisers. In particular, we propose a new approach to combine time-series features and time-invariant features in the deep learning model, to process diverse sources of input data. Compared with baseline models, our model achieves better accuracy and requires a shorter observation window of the time-varying features from the campaign launch to provide robust predictions with high confidence. To extract interpretable insights, we further conduct a multivariate time-series clustering analysis and identify four typical temporal donation patterns. This demonstrates the heterogeneity in the features and how they relate to the fundraising outcome. The prediction model and the interpretable insights can be applied to assist fundraisers with better promoting their fundraising campaigns and can potentially help crowdfunding platforms to provide more timely feedback to all fundraisers. Our proposed framework is also generalizable to other fields where diverse structured and unstructured data are valuable for predictions. 


}%


\KEYWORDS{medical crowdfunding, deep learning, multivariate temporal clustering, crowdfunding patterns}

\maketitle

\section{Introduction}

Medical crowdfunding is one type of donation-based crowdfunding, helping users raise funds to pay medical bills by collecting  donations from many people. In contrast to reward-based crowdfunding, where users contribute to projects with the expectation of receiving rewards in equity, cash or product if the project succeeds, for donation-based crowdfunding, users make donations to cases to help individuals out of financial difficulties without expecting direct financial rewards. Medical crowdfunding has seen rapid growth in recent years. Out of the 72.3 billion USD raised on crowdfunding platforms worldwide in 2017, about half was estimated to come from medical crowdfunding \citep{orbis_2018}. Another recent survey indicates that among all users who contribute to crowdfunding, 68\% have donated to non-reward-based crowdfunding projects to help someone out \citep{smith2016shared}.  The benefits of medical crowdfunding manifest, for example, through funding for medical expenses that medical insurance and other safety-net institutions cannot cover and easing the financial burdens from unexpected illness \citep{jopson_2018,young_scheinberg_2017}. 

Despite the growing importance of medical crowdfunding, 
fundraisers still face significant uncertainty on the fundraising outcome due to the large variations in donations to different crowdfunding cases. 
A large number of cases fail to raise enough money through donations to meet their fundraising goals \citep{berliner_kenworthy_2017}, and fundraisers typically would not know how their cases will perform until they approach the end of the campaigns. This uncertainty in fundraising outcome impacts decisions such as whether to receive a medical treatment, as fundraisers cannot reliably know whether they will collect sufficient funds to cover medical bills afterwards. GoFundMe, one of the leading crowdfunding platforms, estimate that 250,000 medical crowdfunding campaigns are launched each year, with one third involving paying medical bills \citep{mcclanahan_2018}. In countries where pre-payment is required to receive a medical procedure, it is even more critical to provide early predictions about the donations fundraisers are likely to receive. For cases that would eventually fail, early warnings give fundraisers more time to plan ahead and start looking for alternative solutions. Thus, our primary goal in this paper aims to provide a model to provide early and reliable predictions.

Existing work on crowdfunding performance mostly focus on analyzing how time-invariant case attributes influence the outcome \citep{du2015money,greenberg2013crowdfunding}. These attributes, including fundraisers' age, gender, social connections and ability to use online tools have been found to influence crowdfunding success \citep{young_scheinberg_2017, valle_2017}. 
However, time-varying case attributes have not been examined closely in existing studies to explore how they can provide dynamic predictions or updates for the total sum of donations a case will receive. This is due to limitations in data availability and challenges for models to process diverse types of data. 
As daily inflow of detailed information becomes increasingly available, it is imperative and intriguing to examine the temporal behavior of crowdfunding cases. Our second goal in this paper is to develop novel algorithms based on deep learning models to incorporate different types of features, time-varying and time-invariant, to improve the performance of the prediction model. In addition, we aim to identify the latent patterns in the features and obtain proper insights from the daily observations on the fundraising performance.

To achieve these goals, we collaborate with one of the largest medical crowdfunding platforms in China. The platform has helped 630,000 patients raise funds for their medical expenses to date, with the monthly amount of donations averaging around 1.5 million RMB (about 200,000 USD). We collect a unique dataset with detailed case-level information, including both structured data and text data, and dynamic observations for 51,288 unique cases. For each day that case is active, we obtain features such as online communications initiated by fundraisers, including the number of replies sent to potential donors and status updates about the case, and activities initiated by potential donors, including the number of users who verified, shared or made donations to the case. To our knowledge, such detailed case-level data has not been available for use in prior studies.

We design a machine learning model exploiting both the time-varying variables and time-invariant variables to predict the amount of donations cases receive on a daily basis. We use LSTM (Long Short-Term Memory) model to process sequences of time-series features. To combine time-invariant variables and time-series features, we design a new framework where the hidden states of LSTM units are initialized with the time-invariant features. This way, the LSTM is conditioned on time-invariant case-level inputs, mimicking that the daily donations received are conditioned on the case attributes since its launch on day 0. The time-invariant inputs used are concatenations of learned representation of the fundraising posts (text data) and other structured features (age, gender, etc), to incorporate the diverse information input for each case.  
To train the model, we extract six weeks of daily metrics since the launch for each case and obtain a total of 2 million instances. 
We compare the proposed model to baseline deep learning models, including two popular existing approaches of combining time-invariant and time-series features from the literature and find our proposed model has better predictive accuracy. This is due to the advantage of our model in allowing the time-invariant features to affect the output at an early stage of the model and avoid polluting the additional input time steps with redundant, non-temporal information. 
We further examine the robustness of the models, focusing specifically on the \textbf{timeliness}  of the predictions. Timeliness measures how quickly the model can generate with high confidence a satisfying prediction that does not exceed a certain threshold of error rate. This measure incorporates two factors of practical importance in the context of medical crowdfunding:  \emph{wait time} to get an accurate estimation of the donations and  \emph{confidence} of predictions, assessing the prediction performance on individual cases in addition to the aggregate level prediction accuracy.  
  Results show that for predicting with 90\% confidence, our model can provide a forecast with 20\% percentage error within one week while other models need to wait at least another week to obtain enough observations to achieve similar accuracy. 
 
 In addition, we evaluate the social welfare implications for applying our model in reducing the wait time to get a reliable estimation, compared with simply estimating based on experience from observing historical cases. Results show that our model saves about two weeks for predictions with confidence 90\% and four weeks for predictions with 95\% confidence, compared with using model-free experience-based predictions.

We observe that while the average performance of different models are comparable, their timeliness vary significantly, implying large heterogeneity in the time-varying features. Thus, we extract more insights 
through mining temporal patterns of medical crowdfunding cases from time-varying features. We conduct multivariate temporal clustering using K-means clustering for each feature followed by K-modes clustering of cases across different features, in order to identify prototypical behavior in time-varying features. We discover four clusters, which we label as  ``\textbf{low interest},'' ``\textbf{active repliers},'' ``\textbf{social attention attractors}'' and ``\textbf{young and female},'' respectively, based on the temporal patterns in time-invariant characteristics of each cluster.  

This paper makes contributions in the following four areas. \emph{First}, our model can provide fast and reliable predictions of the total sum of donations, which can help stakeholders make early decisions. This is especially valuable when crowdfunding will fail since our model can provide a very early warning to grant raisers more time to search for alternatives. \emph{Second}, we propose methodological innovation in combing diverse types of inputs (time-invariant structured attributes, text data and time-varying attributes) into state-of-the-art machine learning techniques. We also designed new evaluation metrics for medical crowdfunding models, considering the needs of fundraisers and crowdfunding platforms.  \emph{Third}, our provide interpretations on how different types of attributes relate to the fundraising outcome, based on empirical understanding of the latent types of crowdfunding cases and identifying the key attributes that differentiate a cluster from others. These insights shed light on strategies fundraisers can take to attract more donations. \emph{Fourth}, our modeling technique is generalizable to other applications with both time-invariant features and time-varying features, such as predicting stock price, product sales or number of active users.


The rest of the paper is organized as follows. In Section \ref{sec:ref}, we briefly review prior literature on crowdfunding and discuss how our paper differs from them, then we review the state-of-the-art deep learning techniques and highlight the innovations of our model. We discuss the data collection and preliminary data analysis in Section \ref{sec:data}. Then, we present our proposed deep learning model in Section \ref{sec:deep} and examine the performance of the model in Section \ref{sec:exp}.  In Section \ref{sec:timevarying}, we explore the heterogeneity in the time-varying features via clustering analysis to identify different temporal patterns, which further explain results in Section \ref{sec:exp}. 
Finally, we summarize the findings and discuss the managerial insights in Section \ref{sec:con}.

\section{Related Work}\label{sec:ref}
We review the related work on crowdfunding and discuss how our paper contributes to existing literature. Then we discuss related machine learning techniques on deep learning generally and Long Short-Term Memory (LSTM) specifically.
\subsection{Related work on crowdfunding}
Most of the existing studies on crowdfunding focus primarily on reward-based crowdfunding. These studies mainly use explanatory models to explore factors that influence users' decisions to contribute to certain projects, relating to project attributes such as founder gender and ethnicity \citep{younkin2017colorblind}, geographic locations of the founders \citep{agrawal2011geography, lin2015home}, endorsement by other users \citep{bapna2017complementarity}, disclosure of previous contributor information \citep{burtch2016secret} and use of emotional appeals along with quantifiable project quality information \citep{steigenberger2018extending}. While some of the heterogeneity in fundraising success for medical crowdfunding cases come from seemingly similar sources, such as the gender, age and locations of the patients involved, the time-varying features such as social network shares, verifications, and fundraisers' online communications also likely predict fundraising outcomes. How these various attributes interact to influence donation patterns requires an exploration into latent case types. In addition, since medical crowdfunding cases are often time sensitive, using these collections of attributes to provide accurate and timely predictions of funding outcomes is of greater relevance. 


Related studies on crowdfunding also explore the role of social networks and social media on fundraising success. Studies find that social connections (measured by the number of friends on social media) and social network structure (having denser connections in groups of friends) positively relate to funding outcomes \citep{hong2018embeddedness, mollick2014dynamics}. Use of online communications tools, such as posting an update on the status of crowdfunding projects, also leads to better fundraising outcomes \citep{xu2014show}. Most of the existing studies only use snapshots of data instead of including time-series observations on contributions to projects over time.  Extending from this literature stream, our study combines the time-varying features on users' social media activities and online communications to build a prediction model that dynamically predicts the amount of donations in order to provide more timely feedback to fundraisers. 

The emerging stream of literature on medical crowdfunding mostly uses aggregate-level data to study the overall impact of medical crowdfunding on the regional level \citep{burtch_chan_2018} or relies mainly on observational data to analyze factors that influence the credibility of medical crowdfunding campaigns \citep{kim2016power}. Extending from this literature stream, we use detailed case-level observations to provide dynamic predictions on the donations collected by each case on the daily level. To our knowledge, this is the first study to build a machine learning model that directly predict the amount of donation for medical crowdfunding using mixed types of features: time-invariant, time-varying, and text data. 

Relating to crowdfunding predictions, existing studies mainly focus on predicting the success of reward-based crowdfunding, where the problem is cast as a classification problem with success defined as meeting the funding target. They either use traditional machine learning models such as logistic regression \citep{du2015money}, decision tree \citep{greenberg2013crowdfunding}, or topic modeling \citep{yuan2016determinants} to only work with time-invariant features or manually engineer features from time-series and then apply decision trees \citep{rao2014emerging} or K-nearest-neighbor \citep{etter2013launch}. The methods used are conventional machine learning models that work with panel (tabular) data of reasonable size. In contrast, our dataset consists of both time-varying and time-invariant features and a total of two million data points. Both the dimensions of features and the number of observations are beyond the ability of traditional classifiers. 
In addition, previous predictive models make one prediction for each project, when it is just launched or halfway through the campaign. Here, we aim to produce a forecast every day, starting from before a case is launched to give users immediate feedback, and then provide updated predictions on a daily basis as new data comes in, so users are informed on the progress of the case throughout the time the case is active. The complexity of the dataset and the problem exclude the conventional machine learning models from consideration. To the best of our knowledge, this is the first study to use deep learning to make daily predictions on the total sum of donations, using detailed case-level data and daily inflow of time-series.


\subsection{Related work on deep neural networks for time-series}
Deep learning \citep{lecun2015deep} has become one of the most promising and powerful machine learning models in recent years, and it has demonstrated great potential in various fields, including computer vision, speech recognition and natural language processing.

One unique advantage of deep learning models is their ability to  learn hierarchical representations of feature input automatically, in contrast to traditional machine learning models that rely on engineering features manually \citep{farabet2013learning}. A deep neural network (DNN) transforms an input $\mathbf{x}$ through multiple hidden layers with different carefully designed activation functions, such that the output of the model can approximate an output $y$. According to the universal approximation theorem \citep{csaji2001approximation}, a neural network can approximate any continuous function, provided it has at least one hidden layer and uses non-linear activation functions on that layer. This theorem provides a theoretical foundation and explanation for the superior performance of DNN in many predictive tasks, compared to traditional machine learning models. The advantage of DNN is especially salient when the size of the training data is large.

Recurrent Neural Network (RNN) is a type of DNN for analyzing sequential data \citep{graves2013speech}. The loops in the network design of RNNs selectively pass information across sequential steps and allow information to persist in ``memorizing'' temporal patterns in the data. RNN has been widely applied in areas such as language modeling \citep{mikolov2010recurrent} and acoustic modeling \citep{sak2014long}. Among the family of RNN models, Long Short-Term Memory (LSTM) \citep{hochreiter1997long}, which is capable of learning long-term dependencies, is the most popular choice. LSTMs were developed to deal with the exploding and vanishing gradient problems that can occur when training traditional RNNs \citep{hochreiter2001gradient}. The memory cell in LSTM contains a node with a self-connected recurrent edge of fixed weight one, ensuring that the gradient can pass across many time steps without vanishing or exploding \citep{lipton2015critical}. Relative insensitivity to gap length is an advantage of LSTM over RNNs, hidden Markov models, and other sequence learning methods in numerous applications.

A standard LSTM unit is composed of a cell, an input gate, an output gate and a forget gate. An LSTM cell computes the following functions
\begin{align}
 \text{Input gate: }&  & i^{[t]} &= \sigma(W_i\cdot [h^{[t-1]},x^{[t]}] + b_i)\\
 \text{Forget gate: }&   &  f^{[t]} &= \sigma(W_f\cdot [h^{[t-1]},x^{[t]}]+b_f)\\
 \text{Cell state: }&  &  c^{[t]} &= f^{[t]} * c^{[t-1]} + i^{[t]} * \text{tanh}(W_c * [h^{[t-1]},x^{[t]}]+b_c)\\
  \text{Output gate: }&    & o^{[t]} & = \sigma(W_o \cdot [h^{[t-1]}, x^{[t]}]+b_0) \\
  \text{Hidden state: }&  &    h^{[t]} &= o^{[t]} * \text{tanh}(c^{[t]})
\end{align}
where $t$ is the time step in terms of days, $h^{[t]}$ and $c^{[t]}$ represent the hidden state and the cell state at time $t$, respectively. $x^{[t]}$ is the hidden state of the previous layer at time $t$. $i^{[t]}, f^{[t]}$ and $o^{[t]}$ are the input, forget and output gates, respectively. $\sigma$ is a sigmoid function where $\sigma(x) = \frac{1}{1+e^{-x}}.$ 

LSTMs contain information outside the normal flow of the recurrent network in a gated cell. Information can be stored in, written to, or read from a cell, much like data in a computer?s memory. The cell makes decisions about what to store and when to allow reads, writes, and erasures, via gates that open and close. Unlike digital storage on computers, however, these gates are analog, implemented with element-wise multiplication by sigmoids, which are all in the range of 0-1. Analog has the advantage over digital of being differentiable and, therefore, suitable for backpropagation.
The cell remembers values over arbitrary time intervals, and the three gates regulate the flow of information into and out of the cell.


\subsection{Combining time-series with Time-invariant Features}\label{sec:combine}
One major challenge in applying RNN is incorporating information from time-invariant features into the deep learning network together with time-series features. This challenge frequently occurs in healthcare and medical diagnosis where the data involves both static patient information such as age and gender, and temporal features, such as diagnosis, test results and other measures of patient status collected at different time stamps. As new technology is continuously being developed to enable real-time data collection, it is increasingly important for firms in various industries to analyze such time-varying input, in combination with other static input data. 

\begin{figure}
\FIGURE
{\includegraphics[width=0.95\textwidth]{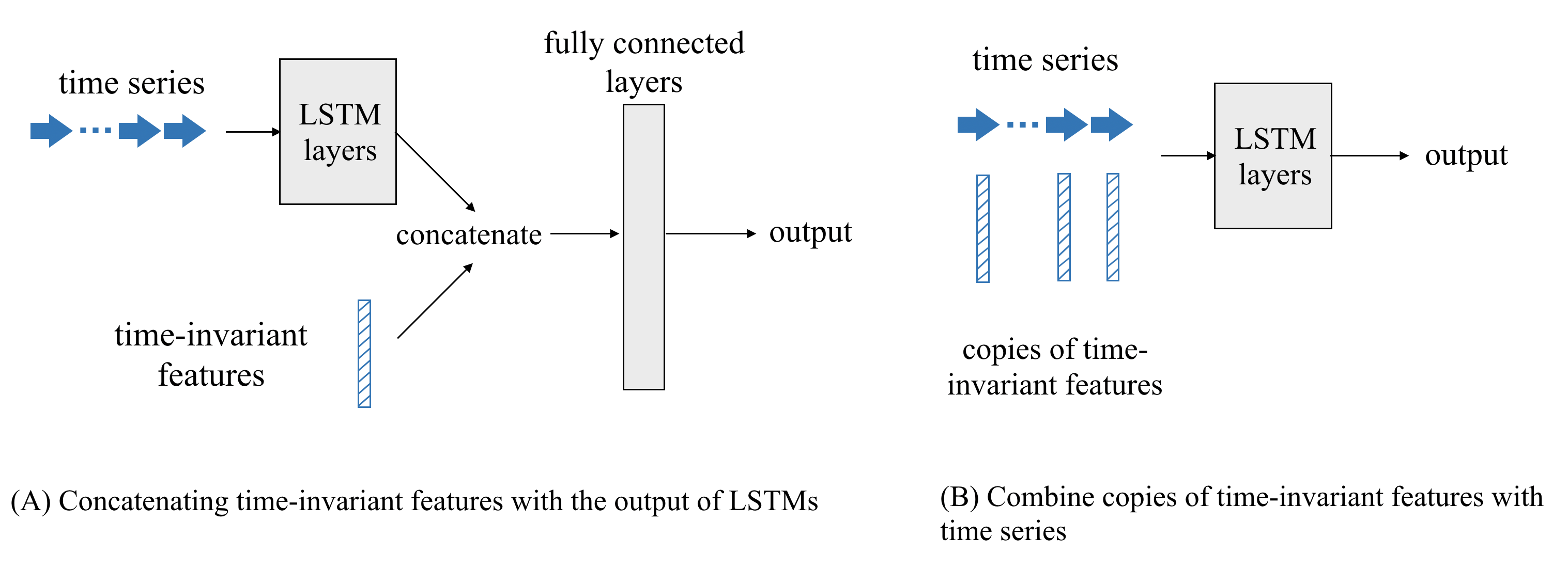}}
{General Frameworks of the Two Existing Approaches for Combining Time-series with Time-invariant Features in a Deep Neural Network Model \label{fig:LSTM_baselines}}
{}
\end{figure}

There are two main approaches in the existing literature to deal with this challenge. The first approach is to feed time-series features to RNN and then concatenate with static features \citep{zhu2018churn,leontjeva2016combining,lin2018early,esteban2016predicting}. For example, \cite{lin2018early} builds a deep learning model to early diagnose sepsis shock by concatenating the static information with the output of LSTM in the final time step. Using similarly structured data, \cite{esteban2016predicting} proposes to concatenate the static information with the hidden states of LSTM at every time step, which improves model accuracy compared with concatenating only at the last step. The second approach for combining the two types of features is to include the time-invariant features as part of the temporal features and feed them together to RNN units \citep{donahue2015long,hsu2019enhanced}. With this approach, the RNN units will see multiple copies of the time-invariant input variables along the sequential steps. Compared to the previous approach, this method involves static information at an early stage of the model, but the downside is that it pollutes the temporal features with non-temporal information. The general frameworks of the two approaches are illustrated in Figure \ref{fig:LSTM_baselines}.

Extending from existing literature, we propose a new design to concatenate the two types of features, attempting to reap the benefits of both main approaches and reduce the flaws in each approach. Our design is motivated by the works on image description generation \citep{karpathy2015deep,vinyals2015show}, using LSTM units to generate text data where the hidden states of the first LSTM layer are initialized with the embedded image representation. This model shows superior performance over baseline LSTM models. We propose to extend this idea and use the embedded time-invariant features, including structured features and texts, to initialize the hidden states ($h^{[t]}$) of the LSTM units in our model. 
This approach ensures that the time-invariant information is incorporated in the RNN units and, at the same time, avoids the information redundancy generated in populating time-series input with repeated copies of time-invariant features.  

\section{Data Collection}\label{sec:data}
We collect a large and unique dataset on fundraising process on a large sample of medical crowdfunding cases, from a leading medical crowdfunding platform in China. The dataset used in this study contains a total of 51,228 cases from October 2016 to June 2018. The data contains both static case attributes and dynamic daily observations on fundraising progress and interactions between fundraisers and potential donors on social media. 

\begin{figure}
\FIGURE
{\includegraphics[width=0.95\textwidth]{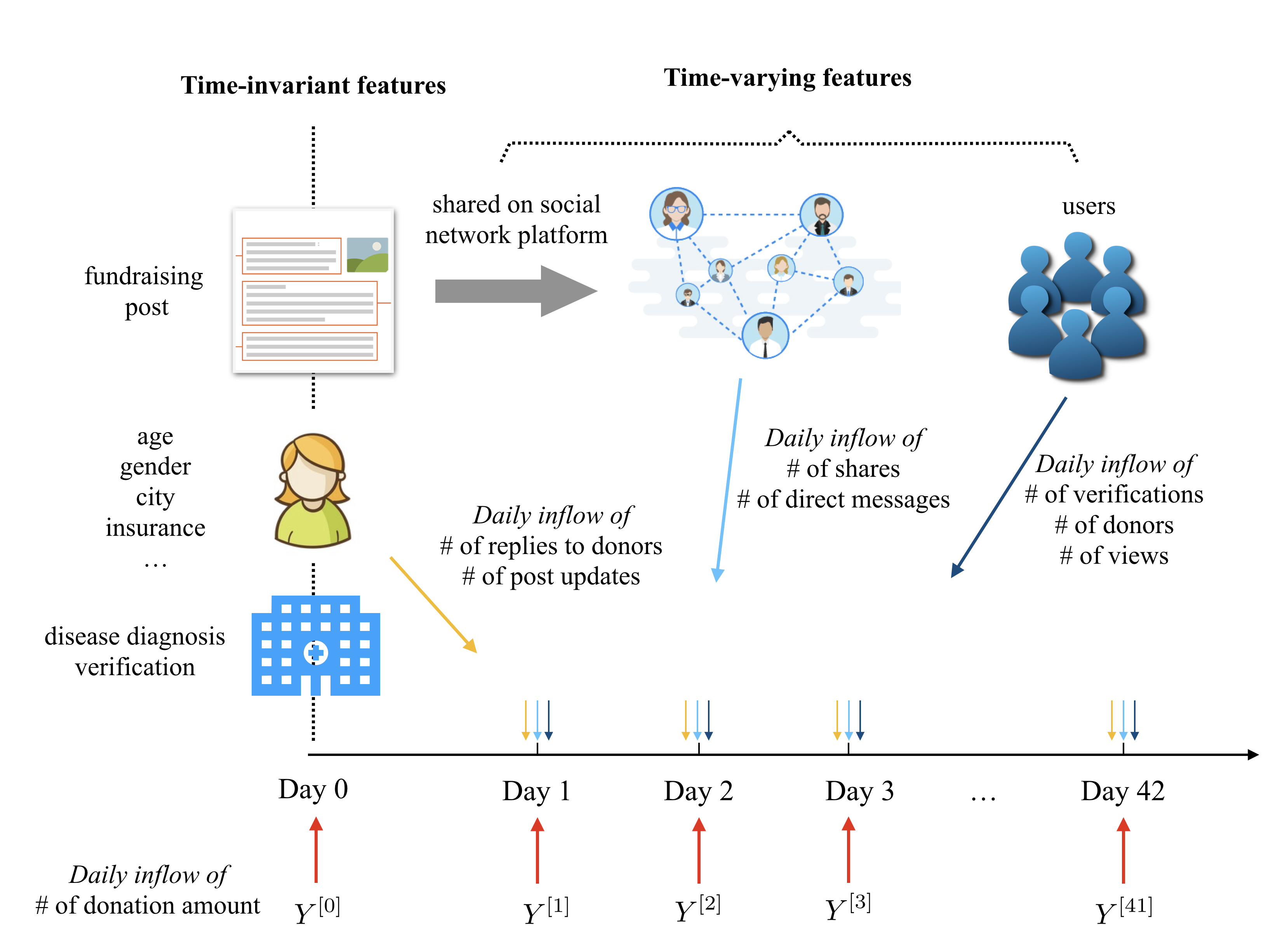}}
{An Overview of the Crowdfunding Process \label{fig:overview}}
{}
\end{figure}

A highlevel overview of the fundraising process on this medical crowdfunding platform is shown in Figure~\ref{fig:overview}. Typically, the fundraiser can either raise the money for himself or on the patient's behalf. When a fundraiser creates a new case on day 0, he submits patient information including demographics (age, gender, location), insurance status (commercial or basic medical insurance) and the target amount, along with a short ``call for donation'' post.  Verification from the patient's hospital is also obtained to prove the validity of the case.  We call these attributes time-invariant features since they will remain the same throughout the crowdfunding campaign. Once the case is reviewed and approved by the platform's employees, it will then be published on the platform.

In addition to the time-invariant attributes, we also obtained daily metrics on case sharing and performance collected by the platform starting from day 0. After a case is published, the link to the page can be distributed through different social network platforms such as Wechat and QQ. Users interested in the case can click the link to browse case information and potentially make donations. Therefore, we collect a total of eight time-series features (see Table \ref{tab:ts}) from the following three main sources: 
First, we collect the number of times a case is shared on the different social networks. Since WeChat is the dominating social media platform in China (the equivalient of Facebook in the US) and covers a more diverse population than other social networks, we separate the number of shares on WeChat moments (which are similar to Facebook posts) from the other social networks as an independent feature. In addition, WeChat has a feature that allows a user to send a case link directly to a friend via Wechat message (which is similar to Facebook direct message), so we also collect the number of times the case is shared via Wechat messages. 
The second source is the fundraiser. We collect the number of times a fundraiser responds to users' questions and requests for more information (or proof) on the case pages. The third source of the temporal information is the crowdfunding platform, which provides records on a daily level, the number of views for the case, the number of users who verified the case, the number of users who donate and the donations amount (measured in both daily donations and total amount donated to the case). 
\begin{table}[]
\centering
\caption{The Temporal Attributes Collected Daily}\label{tab:ts}
\begin{tabular}{cll}
\toprule
\textbf{Index}&\textbf{Temporal Attributes} &\multicolumn{1}{c}{\textbf{Definitions}} \\\hline
1&donate\_amt  & the amount of money received through donations (in RMB)         \\
2&donate\_user\_num  &the number of unique users who donate \\
3&reply\_num   & the number of times a fundraiser replies to users online \\
4&verification\_user\_num   & the number of users who verify the case\\
5&share\_wechat\_msg   & the number of times a case is messaged to a friend via Wechat\\
6&share\_wechat\_moment  & the number of times a case is shared on Wechat \\
7&share\_cnt   & the total number of times a case is shared on social networks         \\
8&total\_view   & the number of times the case is viewed\\ \bottomrule
\end{tabular}
\end{table}

 The platform does not specify a maximum for how many days a case can remain active. Users can continue to view and donate to cases until the fundraisers decide to close the campaign and withdraw the funds. When the campaign closes, the final amount of donations collected can be less or more than the target amount. However, from the preliminary analysis, we observe that 
the average amount of daily donations are close to 0 after six weeks (42 days).  Therefore, we chose the window of observation to be six weeks after the initial posting for all cases. 
The average total money donated is 31,143 RMB from our sample, which matches the average for all cases on the platform\footnote{The unit for all donation amounts is Chinese RMB.}. 
Besides the received donation amount, we are also interested in knowing how much the fundraisers have received with respect to the target goal amount. The mean target amount is 195,676 RMB. The portion of donations actually received compared with the target amount, is what we define as \emph{fulfillment}. 

The majority of fundraisers received a small fraction of the target amount. Only 7.2\% of them reached half of their goal and about two-thirds of the fundraisers only received up to 20\%. The correlation between the target donation amount and the fulfillment is negative, -0.19, which suggests that the higher the target donation amount, the less likely the fundraiser is in reaching his goal.

\subsection{Preliminary Analysis of Time-Invariant Features}
As preliminary study, we examine how each time-invariant feature affects the total amount of money donated. These features are available from day 0, immediately after a fundraiser submits an application to provide relevant information about the case and gets approval to start the fundraising campaign on the platform. The time-invariant variables include the patient's demographic information, disease information, verification from the hospital, the length of the case descriptions post, and the length of the case title are observed. 

\OneAndAHalfSpacedXI 
\begin{table}[]
\centering
\caption{Regression Results of Time-Invariant Case Features on Money Raised Through Donations}
\label{tab:ols}
\begin{tabular}{lll}
\toprule
DV: Log(Total Donations)                        & (1)       & (2)        \\\hline
Log(Target Amount)                              & 0.447***  & 0.419***   \\
                                                & (0.0102)  & (0.0102)   \\
Female Patient                                  & 0.0513*** & 0.0513***  \\
                                                & (0.0178)  & (0.0177)   \\
Patient Age                                     & -0.185*** & -0.193***  \\
                                                & (0.00884) & (0.00891)  \\
Patient Age \textasciicircum 2                  &           & -0.0314*** \\
                                                &           & (0.00533)  \\
Has Basic Medical Insurance                     & 0.168***  & 0.170***   \\
                                                & (0.0197)  & (0.0196)   \\
Has Commercial Insurance                        & 0.0113    & 0.0260     \\
                                                & (0.0425)  & (0.0423)   \\
Length of Case Descriptions                     & 0.323***  & 0.491***   \\
                                                & (0.00883) & (0.0120)   \\
Length of Case Descriptions \textasciicircum{}2 &           & -0.0959*** \\
                                                &           & (0.00450)  \\
Length of Case Titles                           & 0.0732*** & 0.0506***  \\
                                                & (0.00855) & (0.00897)  \\
Length of Case Titles\textasciicircum{}2        &           & -0.0608*** \\
                                                &           & (0.00898)  \\
Fundraiser Gender Disclose                      & 0.249***  & 0.261***   \\
                                                & (0.0386)  & (0.0384)   \\\hline
Regional Controls                               & YES       & YES        \\
Time Controls                                   & YES       & YES        \\
Observations                                    & 51,227    & 51,227     \\\hline
R-squared                                       & 0.422     & 0.428      \\\bottomrule 
\end{tabular}
\centering
\begin{tablenotes}\footnotesize
\item[1] \quad\quad\quad \textbf{Notes}: 1. Time controls include the month, day and day of the week when the campaign is launched 
\item[2] \quad\quad\quad\quad\quad\quad\, 2. Standard errors in parentheses, *** $p<0.01$, ** $p<0.05$, * $p<0.1$
\end{tablenotes}
\end{table}
\DoubleSpacedXI 
\normalsize




Here we use regression models, relating the case-level information to the logged value of the total sum of donations received, and controlling for the location of the patients, the time the fundraising campaign launches (controlling for the month, day and the day-of-the week effects). Results are reported in Table \ref{tab:ols}. We find that posting a larger target amount is associated with larger amounts of donations received. Cases involving female patients receive 5\% larger total donations on average than cases involving male patients. Other case attributes such as younger patients and having basic medical insurance are also associated with larger donations collected. In addition, presenting case information effectively can increase the chances of receiving more donations. For example, the total donation amount received initially increases as the word count for the case description increases and then starts to decrease after exceeding five standard deviations (about 1,700 characters); there is a similar relationship for the word count of the case page title. This suggests that cases with better-designed titles and descriptions that convey information efficiently are likely to see better fundraising performance. On the other hand, the $R^2$ values for these regressions are around 0.4, suggesting that about  40\% of the variance in the data is captured by the model. This implies there are other sources of variance that are not captured by time-invariant features effectively. 

The analysis so far implies that case-level features alone are insufficient of providing an accurate prediction of crowdfunding outcomes. Therefore, it is important to examine the time-varying features of the case as well to evaluate how factors such as social media activities and online communications contribute to the campaign donations and provide actionable insights for fundraisers. 

In the rest of this paper, we build a deep learning model incorporating diverse inputs to provide a reliable prediction as early as possible, utilizing information received up to date, including the daily inflow of various time-series features received until the day of the prediction, the time-invariant structured features and the fundraising post. 

\section{Daily Predictions for Crowdfunding Outcomes}\label{sec:deep}
\begin{figure}
\FIGURE
{\includegraphics[width=0.95\textwidth]{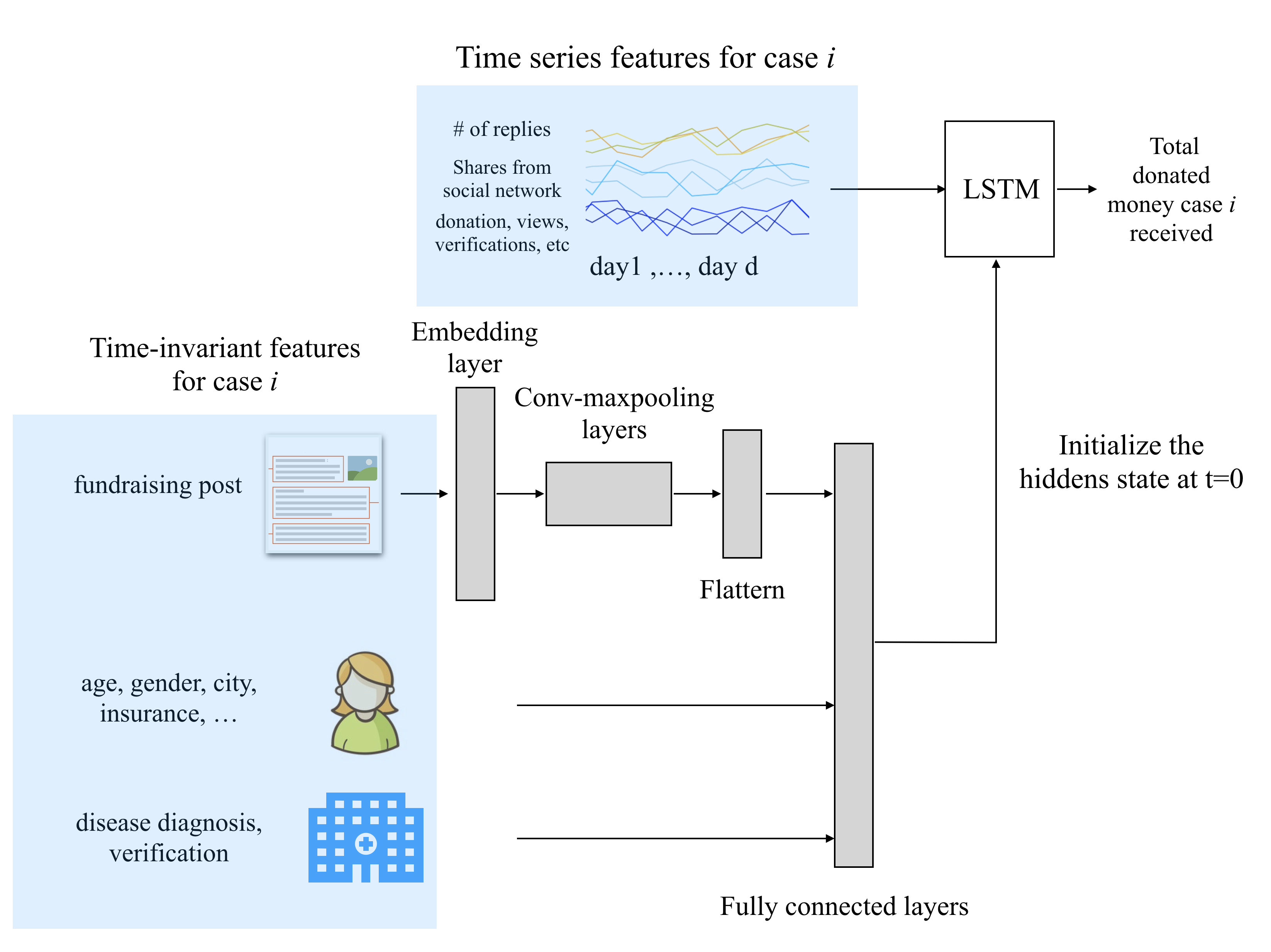}}
{The Architecture of the Proposed Model\label{fig:model}}
{}
\end{figure}

Our dataset contains 51,228 cases. For each case, we collect the time-invariant features and time-series features for the six weeks after the case is launched. We remove observations with missing data points or errors and are left with a total of 48,382 cases. We select a six-week time window since the average daily donations after six weeks approaches 0. With this daily observation of features, we build a model that works with variable sequences of input. When making a prediction on day $d$, the time-series feature contains daily inputs from day 1 to day $d$. Therefore, we got a total of 2 million data points (41 relevant days of observations for each crowdfunding case).\footnote{For each case, we have a total of 42 days of observations from the six-week time window. Prediction on the last day is not relevant since the result is fully known. Therefore, we focus on the time-series predictions for days 1 to 41.} We split the dataset into 80\% training set, 10\% validation set, and 10\% test set. We propose a calibrated framework that combines time-varying features with time-invariant features, including the text descriptions in the fundraising post, to provide an accurate and robust estimation of the donation amount.  
Our model first embeds features extracted from the fundraising posts and then uses the time-invariant features to initialize the states in LSTM units.
Figure \ref{fig:model} provides an overview of the model. We elaborate on each of the components below.

\subsection*{Text Embedding and Feature Extraction}
While the other time-invariant features can be directly applied to initialize the first layer in LSTM, to include information from the fundraising post, we use state-of-the-art designs to extract features from text data. First, we use word embedding to represent the fundraising posts. Word embedding is a class of approaches for representing words and documents using a dense vector representation \citep{levy2014neural}. In this step, the input texts are first integer encoded, i.e. each word is represented by a unique integer, and then each fundraising post is transformed into a matrix, denoted as $\mathbf{d}^e_i$. If we use these representations directly, the matrices are generally large and sparse, making the analysis computationally expensive from the ``curse of dimensionality'' issue. The common practice to address this issue is to use an embedding layer \textbf{emb}$(\cdot)$ that transforms each word into a dense vector which represents the projection of the word into a continuous vector space, i.e.,
\begin{equation}
    \mathbf{d}^e_i=\textbf{emb}(\mathbf{d}_i).
\end{equation}

In the second step, to reduce the size of the learned representation and facilitate extracting meaningful features, we apply a series of convolutional layers and max pooling layers, following an approach that is widely adopted in deep neural nets for text data \citep{lai2015recurrent,dos2014deep,wang2012end}. 
\begin{equation}
    \tilde{\mathbf{d}}_i = \text{MaxPooling-Conv}^H(\cdots \text{MaxPooling-Conv}^1(\mathbf{d}^e_i))
\end{equation}
The number of conv-maxpooling layers, $H$, are tuned via the 10\% validation set. 
Thus $\tilde{\mathbf{d}}_i$ is a learned dense representation of the input fundraising post.
Finally, we concatenate $\tilde{\mathbf{d}}_i$ with the other structured time-invariant features (gender, age, etc.), forming a new vector called \emph{conditions},
\begin{equation}
    \mathbf{c}_i = [\tilde{\mathbf{d}}_i, \mathbf{v}_i].
\end{equation}
Our next goal is to incorporate the conditions into LSTM units and let it affect the output of LSTM.

\subsection*{Conditional LSTM}
We would like the conditions $\mathbf{c}_i$ to affect the model at an early stage, which is consistent with the timeline in the real data: the case-level features are available on day 0 and all daily metrics that happen after that are intrinsically dependent on these features. To incorporate this natural timeline into the model design, we initialize the LSTM with the conditions.
The computation at the hidden states is modified to the following, where at time zero, the hidden state is set to the condition $\mathbf{c}_i$. 
\begin{align}\label{eqn:hidden}
    h^{[t]} &= \mathbbm{1}(t=1)\cdot \text{tanh}(\mathbf{W}_\text{con}*\mathbf{c}_i+b_\text{con}) + \mathbbm{1}(t>1)\cdot  o^{[t]} * \text{tanh}(c^{[t]}),
\end{align}
where $o^{[t]} * \text{tanh}(c^{[t]})$ is the computation of hidden states in a regular LSTM cell, $\mathbf{W}_\text{con}$ and $b_\text{con}$ are trainable parameters in the model. The other operations within a LSTM unit is the same as equations (1) to (4). This formula means that when feeding a sequence of time-varying features, the hidden state for the first time stamp is initialized with a transformation of conditions $\mathbf{c}_i$. This information will stay in LSTM and propagated through future time steps.

An LSTM computes the following functions:
    \begin{equation}
        \mathbf{r}^{[t]} = \text{LSTM}_{k}(\mathbf{e}^{[t]}),
    \end{equation}
$\text{LSTM}(\cdot)$ represents the transformations in the cell as described in equations (1) to (4) and equation (\ref{eqn:hidden}). Using a validation set, we experimented with different layers of LSTMs and set the number of units in the LSTM cell from $[16,32,48, 64,80,96,102,128]$ at each iteration, following the standard approach in training LSTM models.  The best number of layers determined through this search process is 2 and the number of cells is 64 and 32, respectively.
Then, the final output is obtained through transformations at multiple hidden layers: 
\begin{equation}
    \hat{y}_i = \Psi^Z(\cdots \Psi^2(\Psi^1(\mathbf{r}^{[t]})).
\end{equation}
where 
\begin{align}
    \Psi^z(\mathbf{v}_i) &= \text{ReLu}(\mathbf{w}^z \text{Dropout}(\mathbf{r}_k^{[t]} ) + b^h).\label{eqn:psi}
\end{align}
The number of hidden layers, $Z$, is tuned using a validation and set to be 3.  We use ReLu as the activation function and apply 0.1 dropout rate at each layer to avoid overfitting. 


For the training objective, we use the mean squared error, which is a standard loss function for regression problems, to evaluate prediction accuracy: 
\begin{equation}
    l = \frac{1}{N}\sum (y_i - \hat{y}_i)^2.
\end{equation}


\section{Results}\label{sec:exp}
In this section, we report the performance of our proposed  model and compare it with multiple baselines, which either use a subset of features, providing an ablation study for features, or a different model design for combining diverse inputs. We compare the average performance of different methods first and then conduct a robustness check which evaluates how early in the campaign a model can provide a reliable estimation with high confidence. Finally, we evaluate the welfare implications of our model in terms of time saving for patients. We refer to our proposed model as \textit{LSTM-cond} in this section. 

\subsection{Baselines} 
Since our problem is a time-series prediction which takes sequences of data as model inputs and the data contains both structured and unstructured data, traditional machine learning baselines such as linear models or random forests cannot be applied. Thus, we propose different neural network models using different types of features  for the comparison. 
\subsubsection{Models with time-invariant features only}
We build two baseline models using only time-invariant features, \textit{NN: time-invariant (no text)}  and \textit{NN:time-invariant}. The first model uses the structured features and the second model also includes the text information from fundraising posts. These two model are feed-forward, fully connected neural networks (since LSTM models do not apply to data with time-invariant features only). The output of these models are obtained through transformations on the input at multiple hidden layers: 
\begin{equation}
    \hat{y}_i = \Psi^J(\cdots \Psi^2(\Psi^1(\mathbf{v}_i)). \label{eqn:new_psi}
\end{equation}
where the functions of $\Psi^h(\mathbf{v}_i)$ is defined in equation (\ref{eqn:psi}). 
$J$ denotes the number of hidden layers. Similarly, we use ReLu as the activation function and apply 0.1 dropout rate at each layer to avoid overfitting.

To find the best configuration of the neural network model, we follow the standard practice for tuning the number of layers and the number of nodes on each layer using the 10\% validation set. We started with a model with one layer and tune the number of nodes. Using the validation set, we found the best number of nodes is 60. Then we fixed the first layer to have 60 nodes and added a second layer where we repeated the process of tuning. We found the best number of nodes for the second layer was 130. Then we added the third layer and found the best number of nodes was 90. We noticed that there was noticeable gain when adding the second layer, but the improvement was marginal when adding the third. Therefore, we stopped adding layers after the third layer, so that the best model is a neural network with three hidden layers, which obtains a mean absolute error of 18,329 on the test data.

We then build the next model including text information through embeddings and feature extraction. The input to the model is the concatenated conditions $\mathbf{c}_i$, as described in Section \ref{sec:deep}. The rest of the model and the tuning approach are similar to  \textit{NN:time-invariant (no text)}. We refer to this second baseline as \textit{NN:time-invariant}.

\subsubsection{Models with time-series features only}
We first build a baseline LSTM model using only the time-series features. 
We add fully connected layers before the LSTM units for better feature learning and after the LSTM units for fitting the objective function. The number of dense layers, the number of LSTM units and the number of LSTM layers are tuned using the validation set, as described previously. The model is trained by grouping sequences of the same lengths into a batch and looping over all batches with various sequence lengths. We choose the best model by their Mean Absolute Error (MAE) on the validation set. The tuned model consists of one dense layer of 64 nodes, followed by two layers of 64 and 32 LSTM units, respectively, and finally processed by two dense layers of 32 and 16 nodes. We call this model \textit{LSTM-time-series}.

\subsubsection{Models with all features}
Next, we build two baseline models that, just like our proposed model, use all features to make daily predictions but uses the two popular approaches for combining static information with time-series features described in Section \ref{sec:combine}, instead of our proposed approach of intializing the first layer with time-invariant features. We refer to these models as \textit{LSTM-concat} and \textit{LSTM-replicate}, respectively. 
\textit{LSTM-concat} concatenates the output of the LSTMs with case-level features as well as embedded text features. \textit{LSTM-replicate} replicates the time-invariant features multiple times and feeds it into the LSTM, together with the time-series. (Details of these approaches are reported in Figure \ref{fig:LSTM_baselines} in the Related Work section).  
For each model, we tune the number of cells and number of layers using the validation set, with a similar approach as training our proposed \textit{LSTM-cond} model.  

\subsubsection{Models with time-invariant features and daily donation amount}
To show the value of time-series features other than the daily donations received, as an ablation study, we only keep the daily donations variables in the LSTM model, removing the other time-series features in our proposed LSTM-cond model. Thus, the time-series has a feature dimension of one. We call the model \textit{LSTM-cond-partial}. The model structure and tuning procedure is the same as what is described in Section \ref{sec:deep} while using the daily donation amount as the only time-series feature. Table \ref{tab:models} summarizes the inputs used for each model, respectively.
\begin{table}[]
\centering
\caption{A Summary of Models and Their Input Features}\label{tab:models}
\begin{tabular}{l|l|cc|cc}
\hline &\multicolumn{1}{c|}{\multirow{2}{*}{Model}}& \multicolumn{2}{c|}{time-invariant}                                  & \multicolumn{2}{c}{time-varying}                                                     \\ \cline{3-6} &  & \multicolumn{1}{c}{structured data} & \multicolumn{1}{c|}{text data} & \multicolumn{1}{c}{ donation amount} & \multicolumn{1}{c}{other features}   \\ \hline
1&\textbf{LSTM-cond}(proposed)                   & \checkmark                                                  & \checkmark                                            & \checkmark &  \checkmark  \\
2&LSTM-cond-partial              & \checkmark                                                  & \checkmark                                            & \checkmark &    \\
3&LSTM-replicate                     & \checkmark                                                  & \checkmark                                            & \checkmark &   \checkmark \\
4&LSTM-concatenate            & \checkmark                                                  & \checkmark                                            & \checkmark &  \checkmark  \\
5&LSTM-time-series                       &   &                                                        & \checkmark & \checkmark \\
6&NN: time-invariant           & \checkmark                                                  & \checkmark                                            &                           &  \\
7&NN: time-invariant (no text) & \checkmark                                                  &                                                                      &                           & \\\hline 
\end{tabular}
\end{table}

\subsection{Average Performance}
\begin{figure}
\FIGURE
{\includegraphics[width=0.9\textwidth]{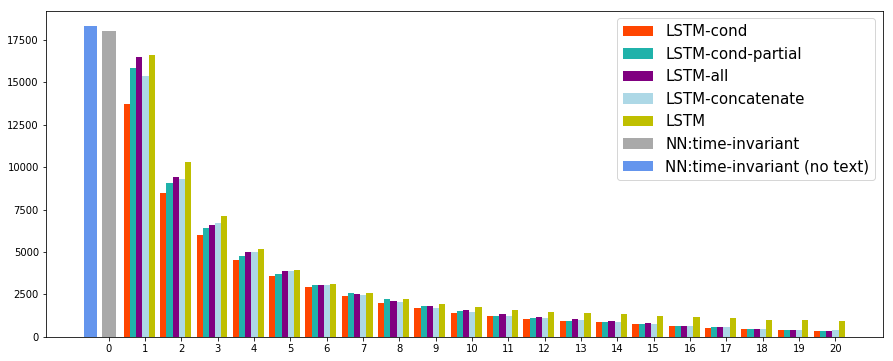}}
{Daily Mean Absolute Error for Predictions Using Different Models \label{fig:mae}}
{}
\end{figure}

We compare the performance of our proposed model (\textit{LSTM-cond}) and the baseline models by comparing the Mean Absolute Error (MAE) for each model. For the two baselines \textit{NN:time-invariant} and \textit{NN:time-invariant(no text)}, which use only time-static features available at the launching of the campaign, we report the model MAE for day 0, since the predictions are not updated and MAE does not change afterwards. For the LSTM models, we plot the MAE for predictions by each model for each day across the entire sample time span of day 1 to day 41 and report it in Figure \ref{fig:mae}. 


The first observation we make from Figure \ref{fig:mae} is that our proposed model (\textit{LSTM-cond}) consistently outperforms all the baseline models. Starting from day 1, our model shows lower MAE than all baselines. This demonstrates that our proposed model extracts pattern from the data for more accurate predictions early on. In addition, while the accuracy for all the LSTM models improve as time elapses and more data points become available, our proposed model continues to outperform the baselines on the subsequent days. 

Comparison with the baselines demonstrates that the better performance of the \textit{LSTM-cond} model comes from the predictive value of both the time-invariant input and the time-series inputs. First, we observe that the MAE for each of the LSTM models has much higher predictive accuracy than using only time-invariant attributes (\textit{NN:time-invariant} and \textit{NN:time-invariant(no text)}). When incorporating time-invariant attributes, we notice that MAE decreases significantly after 20 days. 
On the other hand, the predictions using only time-invariant attributes is not updated over time and does not improve in performance. In addition, the improvement does not simply originate from updates with the daily realizations of donations received. We observe that the \textit{LSTM-cond} model provides more accurate predictions than \textit{LSTM-cond-partial}, which only incorporates the daily donation amounts among the available time-series features. This indicates that the dynamic time-series observations, such as fundraisers' engagement with potential donors on social media, and user activities, such as sharing and providing verification for the case, are useful predictors for the total sum of donations the case is likely to receive. 

Second, we observe that the model using only time-varying features but not the static features (\textit{LSTM-time-series}) performs worse than using both types of features. This finding shows that the time-invariant features reflect information on the heterogeneity in fundraising outcomes and further demonstrates the importance of combing time-varying and time-invariant features in predictive models. In addition, the \textit{NN:time-invariant} model has higher predictive accuracy than the \textit{NN:time-invariant(no text)} model, demonstrating that including information extracted from the text descriptions for each case contributes to improving model performance.  

Furthermore, the \textit{LSTM-cond} model performs better than two alternative approaches of combining time-invariant features and time-series features, \textit{LSTM-concatenate} and \textit{LSTM-replicate}. Particularly, for predictions in the early days of the campaign, the improvement in prediction accuracy is more significant (for example, on day 1, MAE of the \textit{LSTM-cond} model is 10.7\% lower than the \textit{LSTM-concatenate} model and 16.8\% lower than the \textit{LSTM-replicate} model). These results suggest that our approach of initializing LSTM with the static features allows the model to learn patterns from the data more efficiently early on. Intuitively, this is because initializing the LSTM using time-invariant case-level inputs captures the fact that the daily donations received are conditioned on the case launched on day 0 and more accurately reflect how various features of the case contributes to the total donations it will likely receive. 


We have established that our LSTM model shows better overall performance, considering the heteogeneity in fundraising performance across cases (discussed in Section \ref{sec:cluster}) and the time-sensitive nature of medical crowdfunding in practice. We want to explore further whether the model can provide a reliable prediction as early as possible for almost all cases. Therefore, we conduct a robustness check on whether the model provides early forecasts with a high \emph{confidence level} in the next section. 


\subsection{Robustness Checks}
\label{sec:robustness}
Since the purpose of crowdfunding on the platform is financing medical treatments, getting an earlier heads-up on the donations a fundraiser can receive is of paramount importance. If the patient can be informed as early as possible that the donation cannot meet his need, he will have more time to search for other resources, which can be lifesaving in some situations. Therefore, another important criteria for model performance is how soon the model can obtain a reliable estimate of the fundraising outcome. This is represented by how long of an observation window the model requires after campaign launch to predict the donation amount accurately with high confidence. In this section, we examine the robustness of our main model on this attribute and compare it with the same set of baseline models introduced previously.

\subsection{Timeliness Evaluation - Error with Confidence}\label{sec:timeliness}
For most regression problems across domains, average-based metrics such as mean absolute error (MAE) or mean squared error (MSE) are often used. They represent the performance on average but are insufficient to assess performances at the individual level. For example, predictions with bounded error for each crowdfunding case may have the same MAE as predictions that are extremely good for some cases but extremely bad for others. For the second situation, the model could provide totally opposite implications that may mislead fundraisers, such as overestimating the donation amount by a significant amount, so the fundraiser does not look for other financial channels.



To evaluate the confidence of the model predictions, we construct the following measure: for each day $d$ from 1 to 41, we find the smallest percentage error the model can achieve while satisfying a confidence $c$, denoted as $\epsilon^{[d]}_c$. This indicates that with confidence level $c$, the prediction for any case made on day $d$ by the model has an percentage error rate smaller than $\epsilon^{[d]}_c$. This measure is defined as follows:
\begin{equation}
\epsilon^{[d]}_c = \min\{\epsilon | p(\delta^{[d]}_i<\epsilon)>c\},
\end{equation}
where $\delta^{[d]}_i$ is the absolute percentage error in the prediction for case $i$ on day $d$, defined as the absolute error normalized by the true outcome ${y}_i$,
\begin{equation}
    \delta^{[d]}_i = \frac{|y_i - \hat{y}_i^{[d]}|}{y_i}.
\end{equation}
$\hat{y}_i^{[d]}$ is the prediction made on day $d$.
For example, if $\epsilon^{[d]}_c= 10\%$ with confidence $c= 95\%$ and  day $d=7$ where $y_i$ is 10,000 indicates that with 95\% confidence, the model predicts that the total donation amount ]the case receives by day 7 is between 9,000 and 11,000 RMB, within 10\% away from the true value.

\begin{figure}[h]
\FIGURE
{\includegraphics[width=0.95\textwidth]{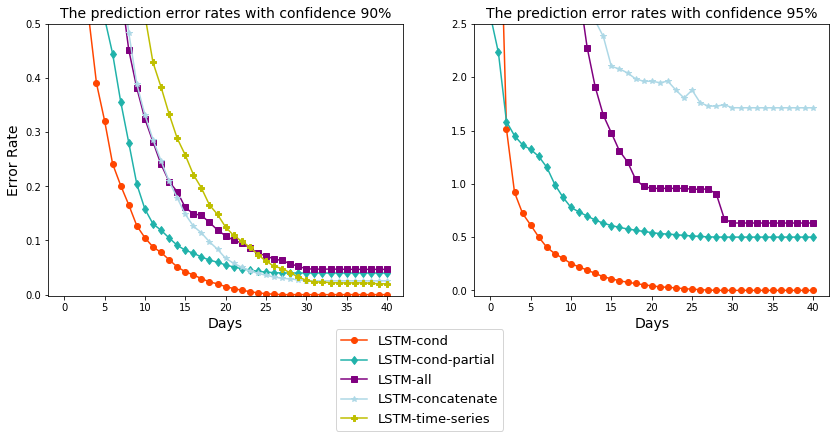}}
{Timeliness of the Predictions  \footnote{Note that the LSTM-time-series curve does not appear in the right figure because the errors are beyond the limits of the current y-axis.}\label{fig:timeliness}}
{}
\end{figure}

To evaluate $\epsilon^{[d]}_c$ on the test set, we compute the empirical probability,
\begin{equation}
    \hat{p}(\delta^{[d]}_i<\epsilon ) = \frac{1}{n}\sum_{i=1}^n \mathbbm{1}(\delta^{[d]}_i \leq \epsilon),
\end{equation}
where $\mathbbm{1}(\cdot)$ is an indicator function which is 1 if the condition holds and 0 otherwise. $n$ is the number of test cases. $\hat{p}(\delta^{[d]}_i<\epsilon )$ represents the percentage of test cases with the absolute percentage error less than $\epsilon$ when predicted on day $d$. 
We set $c$ to 90\% and 95\% and plot $\epsilon^{[d]}_c$ at different $d$ for each model in Figure \ref{fig:timeliness}.

We observe that each model provides more accurate predictions (lower possible error rate for both the two given confidence levels) when longer time windows of data are available. Comparing different model designs, our proposed model \textit{LSTM-cond} requires a shorter time to achieve the same possible error rate compared with the baseline models. In the case of satisfying 90\% confidence, \textit{LSTM-cond} requires seven days shorter time window than \textit{LSTM-replicate} and \textit{LSTM-concatenate} for different values of the error rate. In other words, \textit{LSTM-cond} reduces the time required by at least one week to provide predictions of similar accuracy as other baseline models. In the case of satisfying 95\% confidence, \textit{LSTM-cond} is faster by more than two weeks. 

The comparison between \textit{LSTM-cond} and \textit{LSTM-cond-partial} demonstrates the value of time-varying features besides the daily donation amount. \textit{LSTM-cond} using all time-series features performs better than \textit{LSTM-cond-partial} in terms of requiring shorter observation time to provide the same predictive accuracy. In addition, the performance of \emph{LSTM-time-series}  (using only time-varying features) demonstrates the value of time-invariant features. With 90\% confidence, \emph{LSTM-time-series}) are two weeks behind other baselines also using time-invariant features and at 95\% confidence, \emph{LSTM-time-series}'s performance downgrades so severely that it is not visible in Figure \ref{fig:timeliness} anymore (out of the current Y limits). 

\textbf{Implied Heterogeneity in Time-Varying Features} Comparing Figure \ref{fig:mae} and Figure \ref{fig:timeliness}, we notice that while the average performance (see Figure \ref{fig:mae}) of different models in Table \ref{tab:models} are comparable, the timeliness vary significantly in Figure \ref{fig:timeliness}. This implies large potential heterogeneity and local patterns in time-varying features. We will conduct further analysis on time-varying features in Section \ref{sec:timevarying} to explain this observation.



\subsubsection{Economic and Welfare Implications}
The calculations for $\epsilon^{[d]}_c$  and the illustrations in Figure \ref{fig:timeliness} allow us to find out the number of observation days our model needs to wait in order to get a reliable estimate within $\epsilon^{[d]}_c$ error range from the true amount with confidence $c$. One potential concern about the value of the prediction model in practice is that fundraisers could also observe the donations received from previous cases and estimate their own fundraising outcomes without applying prediction models. To illustrate the contribution of our model compared with such model-free, experience-based estimates, we compute the time saved by applying our model, from the time required to estimate total amounts of funding for the same value range, without applying the prediction model. Similar evaluations have been performed, for example, in \cite{crowds2019} to show the welfare improvement from implementing machine learning algorithms compared with relying on human expertise.

Using a similar logic, we define the time required without applying the prediction model to obtain estimations of total amount of funding that are $\gamma$ (in percentage) away from the true amount with confidence $c$. We refer to this as the \emph{natural wait}. For example, 
if among all cases that eventually receive 10,000 RMB, 90\% of them receive 8,000 RMB (at most 20\% difference from the actual total amount) by day 10, then the natural wait without using prediction models is 10 days for 20\% percentage error with confidence 90\%.  The natural wait for percentage error $\gamma$ with confidence $c$ is
\begin{equation}
    \tilde{d} = \min\{d|\frac{1}{n}\sum_{i=1} \mathbbm{1}(\frac{\sum_{i}^d m_i}{\sum_{i} m_i}\geq 1 - \gamma) \geq c\}
\end{equation}
Thus, we can compare $\tilde{d}$ with the number of days, $d$, our model needs to wait to get $\epsilon^{[d]}_c = \gamma$. Again, we experiment with the confidence $c = 0.9$ and $c = 0.95$ and choose $\gamma$ from $[0.5,0.4,0.3,0.2,0.1,0.05]$. The observation window will increase as a user demands a lower and lower error rate at a given confidence. We report the results in Table \ref{tab:wait}.

\begin{table}[]
\centering
\caption{The natural wait and the observation window of our model }\label{tab:wait}
\begin{tabular}{c|ccc|ccc}
\toprule
\multirow{2}{*}{$\gamma$ = $\epsilon^{[d]}_c$} & \multicolumn{3}{c|}{Confidence = 90\%} & \multicolumn{3}{c}{Confidence = 95\%} \\ \cline{2-7}
                                               & Natural     & LSTM-cond     & Saved    & Natural     & LSTM-cond     & Saved    \\ \hline
50\%                                            & 15          &   3            &    12     & 42          &  6             &    36     \\
40\%                                           & 17          &   4            &   13      & 42          &  7             &   35      \\
30\%                                            & 19          &  5             &  14       & 42          &      9         &   33      \\
20\%                                            & 22          &   7            &  15       & 42          &     11          &   31      \\
10\%                                            & 25          &  10             &    15     & 42          &    15           &   27      \\
5\%                                           & 28          &   14            &    14     & 42          &   20            &  22    \\ \bottomrule  
\end{tabular}
\end{table}

We observe that to obtain predictions with 90\% confidence, our model requires two weeks less than the natural wait. Across different specifications of prediction error caps,  our model requires only one third to half of the natural wait time on average. The benefit of our model is more significant when requiring more accurate predictions, with 95\% confidence. We find that it is statistically impossible for fundraisers to get such accurate predictions simply based on observing past performance. This is because considerable heterogeneity exists in the donation patterns and large standard deviations in donations on different days; thus, it is difficult to obtain a highly confident estimate. On the contrary, our model can characterize the variations in the pattern and is three to five weeks quicker to provide a reliable estimation.

\section{Retrospective Mining of Patterns in Time-Varying  Features}\label{sec:timevarying}
We discovered that in our proposed prediction model, time-varying features are critical in achieving accurate and early predictions. In addition, results from Section \ref{sec:timeliness} suggests potential heterogeneity in time-varying features. In this section, we further interpret how each feature relates to the predicted outcome and explore whether there are typical temporal patterns that are associated with better fundraising performance.

\subsection{The Heterogeneity of Time-Varying Features}\label{sec:cluster}
To identify the heterogeneity in donation patterns both across cases and across time, we conduct clustering based on the time-series features to partition cases into groups with similar temporal patterns. For each case, we introduce a $k$-means clustering step to discover the representative patterns for each feature first and subsequently use the $k$-mode method to cluster the cases based on each feature's cluster labels. This clustering approach allows us to group cases into similar clusters based on representative time-series patterns in attributes and is more efficient than using the raw values for each feature in grouping cases directly.

\begin{figure}
\FIGURE
{\includegraphics[width=0.45\textwidth]{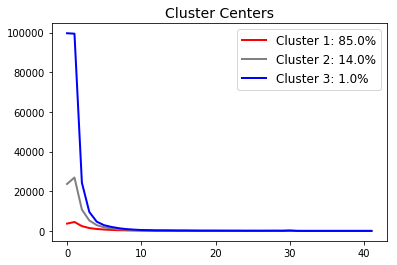}}
{Cluster centers and cluster sizes for attribute ``donate\_amt''.\label{fig:donate_cluster}}
{}
\end{figure}

\paragraph{Step 1: Learning temporal patterns in single features} To identify patterns in each feature as the first step, we apply the $k$-means clustering with Silhouette analysis, following a similar approach as first proposed in  \cite{rousseeuw1987silhouettes}, to each time-series feature $j$. The time-series observations for case $i$, feature $j$ are denoted as  $\{s^{[t]}_{i,j}\}_{t=1}^{42}$. 
This clustering method automatically decides the number of clusters $k$ while partitioning data into $k$ clusters. When performing clustering analysis for feature $j$, each instance is represented as a vector of $42$ elements, corresponding to the values of feature $j$ from day 1 to day $42$. For example, for the feature ``donate\_amt'' ($j=1$),$\{s^{[t]}_{i,1}\}^{[42]}_{t=1} = \{3500,6000,\cdots\}$ represents that the fundraiser receives 3,500 RMB on day 1, 6,000 RMB on day 2, etc. Then, for case $i$, we obtain its cluster label $L_{i,1}$. We obtained three clusters when clustering donation amount, so $L_{i,1} \in \{1,2,3\}$. We plot the cluster centers in Figure \ref{fig:donate_cluster}, which are daily donation amount averages for each cluster of cases. Cluster 1 represents a group of cases that received almost no donations, which accounts for about 85\% of all cases. Cluster 2 receives more donations than Cluster 1, with the peak on day 1 and quickly decreases, accounting for about 14\% of all cases. Cluster 3 receives significantly more donations, receiving 100,000 RMB each day in the first two days and decays to almost 0 within one week. 

Following the same steps, we apply this $k$-means clustering method to all eight time-series features so that each crowdfunding case is represented by a vector of eight cluster labels, denoted as $\{L_{i,1},\cdots,L_{i,8} \}$. We call clusters discovered at this step \textit{feature clusters} and their centers \textit{feature centers}. Note that the number of clusters could be different for each feature, depending on the output of the $k$-means model.  

\paragraph{Step 2: Learning temporal patterns for cases} Next, we perform clustering on the case level, based on the cluster-label representations for the eight available time-varying features. 
 We apply the $K$-mode clustering approach \citep{cao2009new}, which is a classic clustering method that works with categorical features.  
We call clusters discovered at this step \textit{case clusters} and their centers \textit{case centers}. A case center is the most prevalent feature combination in that cluster. Here the ``feature'' refers to the cluster labels $L_{i,j}$ from the previous step of feature clustering.  For example, if a case cluster center is $\{2,3,\cdots\}$, it means the first feature (donation amount) is from the 2nd feature cluster for feature 1, the second feature is assigned to the 3rd feature cluster for feature 2, etc. Using the elbow method \citep{bholowalia2014ebk}, we determine the optimal number of clusters is four and, therefore, came up with four distinctive clusters of cases.

\subsection{Cluster Interpretations}
\begin{figure}
\FIGURE
{\includegraphics[width=\textwidth]{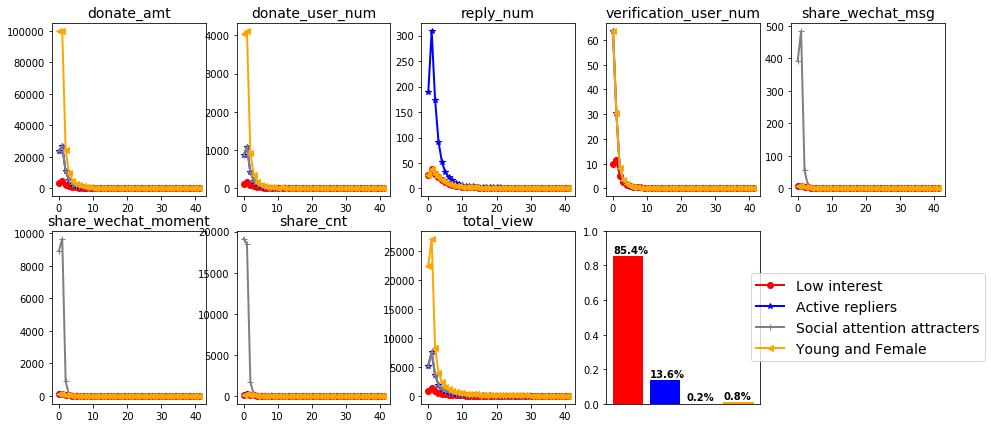}}
{$K$-means Cluster Centers for Different Features \label{fig:cluster}}
{}
\end{figure}

To better visualize and interpret the four clusters, we plot the cluster centers in Figure \ref{fig:cluster}. A cluster center is the most frequent case type, represented by the eight feature centers associated with the type.  We report the mean values of the time-series features and time-invariant features for each cluster in Table \ref{tab:cluster-features}.  Each cluster exhibits different temporal patterns in different time-series features, which we discuss in detail in the following paragraphs.

\begin{table}[]
\caption{Means of Time-series Features and Select Time-invariant Features for Each $k$-mode Cluster}
\label{tab:cluster-features}
\small
\begin{tabular}{l|l|ccccc}
\toprule
\multicolumn{2}{c|}{\textbf{Features}} & 
\textbf{All} & \begin{tabular}[c]{@{}c@{}}\textbf{Low}\\\textbf{Interest}\end{tabular} & \begin{tabular}[c]{@{}c@{}}\textbf{Active}\\\textbf{Repliers}\end{tabular} & \begin{tabular}[c]{@{}c@{}}\textbf{Social Attention}\\\textbf{Attracters}\end{tabular} & \begin{tabular}[c]{@{}c@{}}\textbf{Young \&}\\\textbf{Female}\end{tabular}\\ \hline
\multirow{8}{*}{\begin{tabular}[c]{@{}l@{}}Time-\\ Series \\ Features\end{tabular}}                         
                & donate\_amt                       & 26999     & 17677 & 74782 & 65069  & 246493        \\
                & donate\_user\_num                 & 934       & 574   & 2721  & 2555   & 10028         \\
                & reply\_num                        & 439       & 334   & 1034  & 827    & 967           \\
                & verification\_user\_num           & 50        & 38    & 114   & 110    & 220           \\
                & share\_wechat\_msg                & 33        & 26    & 68    & 811    & 160           \\
                & share\_wechat\_moment             & 478       & 298   & 1155  & 21252  & 4595          \\
                & share\_cnt                        & 808       & 515   & 1809  & 43298  & 7645          \\
                & total\_view                       & 9356      & 6326  & 25417 & 25160  & 75560         \\ \hline
\multicolumn{1}{c|}{\multirow{5}{*}{\begin{tabular}[c]{@{}c@{}}Time-\\ Invariant \\Features\end{tabular}}} 
& Age                     & 40        & 41           & 31        & 28     & 22            \\
\multicolumn{1}{c|}{}  
& \% of female            & 40.0\%    & 40.1\%       & 38.8\%     & 43.0\%     & 48.5\%        \\
\multicolumn{1}{c|}{}                                                                          
& Target Amount           & 196,201   & 182,187      & 269,570    & 268,861     & 383,299       \\
\multicolumn{1}{c|}{}  
& Content Length          & 399 words & 375 words    & 527 words  & 482 words  & 687 words     \\
\multicolumn{1}{c|}{} 
& Title Length           &17 words    &17 words      &18 words     & 19 words & 18 words \\ \hline
\multicolumn{2}{c|}{Fulfillment}& 14.9\% &14.9\%&32.8\%  &28.5\% &69.2\%  \\
\bottomrule
\end{tabular}
\end{table}

Cluster 1 represents a group of cases that received very low interest overall, accounting for 85.4\% of cases, but with the lowest curves and smallest numbers for all time-series features. From the summary statistics for each cluster reported in Table \ref{tab:cluster-features}, we observe that on average, this cluster of cases receives 17,677 RMB in donations from about 574 donors, which is only 20\% that of the cluster with second fewest donors. The average age of the patients involved for this group is 41, the highest among the clusters. This cluster also has the lowest visibility and publicity on social networks, with the least active online case management and response from fundraisers. In addition, despite having lower fundraising goals, this cluster receives the lowest amount of donations and has the lowest fulfillment rate. The fulfillment rate of donation target is also lowest, at about 14.9\% on average, despite aiming for lower target amounts. Half of the cases did not reach even 10\% of their target.  Based on these attributes, we label this cluster as the ``low-interest'' cluster. 

Cluster 2 and 3 are both considered ``medium interest,'' with better overall performance on the eight daily metrics than the ``low-interest" cluster. Interestingly, we observe that the two groups of fundraisers apply different strategies to improve their total donation amount. 
Cluster 2 fundraisers make significantly more communication efforts by actively replying to other users' messages on their case page (with 1,034 replies on average compared with 397 for the other clusters). On the other hand, cluster 3 fundraisers leverage social connections in their fundraising efforts, and their cases are the most widely disseminated on social networks among all clusters (with a total of 43,298 shares on average, more than 50 times the number for the other clusters.)  We refer to cluster 2 as ``active repliers'' and cluster 3 as ``social attention attractors.'' We also observe that the patients involved in cluster 2 and cluster 3 cases are 10 years younger on average than those in cluster 1, which may partially explain the better use of platform communication and social media tools to receive more donations. 

Finally, cases in cluster 4 receive the highest amount of money from the largest number of donors. Interestingly, fundraisers did not make as much effort as those in cluster 2 in replying to messages on the case pages, but they still naturally get a large number of verifications from users and the highest number of views. These cases are not as widely spread on the social networks as cluster 3, but still receives large amounts of donations. By examining the features, we identify possible key factors that determine the success of this cluster, the age and gender. 
From Table \ref{tab:cluster-features}, we observe that cluster 4 is the youngest group, with an average age of 22 for the patients involved. 
It also contains the highest percentage of female patients. This combination of age and gender seems to receive a lot more attention and sympathy than the other clusters. 
We label this cluster  ``young and female'', and it is the most successful cluster. Despite having the highest target amount on average, this cluster achieved significantly higher fulfillment, four times higher than the low-interest cluster and double that for the two medium-interest clusters.

Overall, we observe that part of the cluster analysis results are consistent with the previous regression results in Table \ref{tab:ols}. For example, cases in the ``young and female'' cluster involve more female patients and patients of younger ages,  higher target amount and richer text descriptions on the case details page, which are consistent with findings from the regression analysis. On the other hand, the cluster analysis is able to detect more local patterns that apply to subsets of cases. The cluster analysis uses dynamic features from page views, shares on social media, and fundraisers' communication to provide finer distinctions across the cases, better representing the heterogeneity across cases and across time.   

In addition, the patterns explain the results in Figure \ref{fig:timeliness} in the previous section. With 90\% confidence, \emph{LSTM-time-series} (using only time-varying features) are two weeks behind other baselines while at 95\% confidence, \emph{LSTM-time-series}'s performance downgrades severely.  This is likely because the majority of the users (85.4\%) are in cluster 1 (Section \ref{sec:cluster}), so the time-varying features behave similarly. However, for 95\% confidence, multiple clusters will be involved, thus making the patterns difficult to learn without the information from time-invariant features. Most importantly, the discovered heterogeneity and patterns in time-varying features explain the large disparity in timeliness in spite of their comparable average performances. The average performance is comparable because cluster 1 dominates, so the average daily donations follow a decaying curve that is easy to learn by all models using time-varying features. However, once high confidence is desired, there's a clear demonstration of the powerfulness of some models than others. 
Therefore, it is necessary to take the heterogeneity in donations across cases into consideration and better capture different temporal patterns in our modeling

\section{Conclusion}\label{sec:con}
Despite the fact that more and more people are starting to use crowdfunding to raise money for various purposes, little is done to analyze how different features, especially time-varying features, affect the outcome.  Besides, there has not been any model to predict the amount of money users will eventually receive or the length of time and amount of data needed to get a reliable estimation. Our study provides meaningful implications for both researchers and practitioners.

Most importantly, our study presents a potential solution to the uncertainty fundraisers face in medical crowdfunding through a novel algorithm design based on the state-of-the-art machine learning approaches, which provides accurate and timely predictions on fundraising performance and real-time suggestions about fundraisers' online activities to promote their campaigns. We enrich the literature on medical crowdfunding by examining the dynamic patterns in time-series features, including online communications and the number of shares on social networks. In particular, our proposed approach of incorporating time-series attributes with time-invariant attributes in the deep learning model improves accuracy and shortens the time required to obtain enough observations. Our study shows that our proposed model using time-invariant features to initialize the hidden states of LSTM units, then feeding in the  time-varying features provide the best model for the total sum of donations. Waiting only two days after the campaign launch to obtain time-varying measures, the mean absolute error can be decreased by more than a half than that of the immediate evaluation on day 0.  We then verify the predictive performance using various metrics. Specifically, we test the timeliness of our predictions. Results show that our model needs less than a two-week observation window to predict with less than 10\% absolute percentage error with 90\% confidence.  Compared to waiting without using any models, our results show that our model can significantly shorten the wait time, by two weeks if a user desires 90\% prediction confidence and three to five weeks for 95\% confidence. 

 Through our additional analysis to extract temporal patterns and interpret how they relate to funding outcome, we discover high levels of heterogeneity across cases and identify four clusters of typical temporal donation patterns. Interestingly, we find the majority of cases fall into the ``low interest'' cluster, with low levels of attention from potential donors and donations. Fundraisers' actions such as actively replying to messages on the case page or promoting shares of their case information on social networks, can increase user interest and donations to their cases, moving the cases to the two clusters with middle levels of donations. At the same time, consistent with previous studies, we find a few cases with certain attributes, such as involving young patients or female patients, capture donors' attention naturally, generating significantly higher numbers of shares on social media and donation amounts. Our study not only deepens the understanding of dynamic crowdfunding patterns but also provides a useful instrument for future research exploring latent crowdfunding types. In addition, the analysis on latent case types also complements the prediction analysis, providing an explanation for the comparable MAE across models but large contrast in the confidence measures: that the heteogeneity across temporal patterns is better captured by our proposed model, compared with the baseline models. 
 
For future research, there are a few interesting topics worth pursuing. First, we can potentially enhance the prediction model by adding a richer set of features from image data. Many fundraisers upload photos of patients to attract attention and arouse sympathy. Such additional attributes can be incorporated into the current model by embedding convolutional neurons and potentially improve the accuracy of the predictions. Second, we can conduct a field experiment by deploying our model with the medical crowdfunding platform to provide fundraisers with real-time prediction results. It would be interesting to see how the fundraisers change their engagement behavior based on the predictions throughout the campaigns and how the donation outcomes change accordingly. Moreover, we can study the platform's interventions for cases based on the model predictions, by promoting selected cases or pushing for more shares on social networks, to better allocate resources across different cases. 

Our study has a number of managerial implications. 
From the fundraiser's perspective, our model can help them be better informed about the case's performance throughout the campaign. 
After a case is launched, our model can provide timely predictions of the total donation amount. Particularly, our model gives an early accurate forecast of the donation amount. This information is particularly helpful to fundraisers in evaluating the largest amount of money they are likely to receive by the end of the campaign. Within a few days of launching the campaign, fundraisers can use this evaluation to decide whether they need to promote the campaigns better or find alternative sources of funding if the current predictions are below their goals. In addition, the model can also provide timely campaign managing suggestions for fundraisers to actively manage their campaigns, such as replying more actively to users online, getting more users to verify the cases or sending the cases to more friends on social networks, so it gets disseminated to a larger population.

From the crowdfunding platform's perspective, our findings can help the platform provide more customized suggestions to fundraisers to promote their cases, depending on specific case attributes. Currently, most medical crowdfunding platforms use standardized guidelines for all cases, focusing primarily on supervising the cases on submitting required documentation and providing accurate information on the case details pages. Our model predictions and cluster analysis results can help the platform better differentiate the cases, potentially identifying cases that lack the attributes to immediately attract potential donors attention. Our analysis can help these cases by providing timely feedback for projected donation amounts and suggestions on actions the fundraisers can take to improve case performance. Such an approach provides potential solutions for overcoming the existing disadvantages of certain cases face, such as lacking ``seemingly interesting'' descriptions or lacking rich social connections to disseminate case information on social networks effectively. Particularly, for cases with low levels of attention, we identify actively replying to messages and promoting sharing on social networks as two potential measures to attract more donations to these cases. Applications of our findings can contribute to attracting more donations to cases that were previously overlooked and lead to a more efficient allocation of resources on medical crowdfunding platforms.  

Finally, from the methodological perspective, our model can be generalized to other contexts involving data predictions, where structured and unstructured (text) time-invariant attributes and time-varying attributes both influence the target variables. Such applications include, for example, predicting stock market prices, daily product sales, daily website views and advertisement revenue. Compared with baseline models using only time-invariant attributes, or adding time-varying attributes on aggregate levels, or packaging time-invariant features as time-varying features, our modeling approach provides more accurate predictions overall and is more time-efficient.

\bibliographystyle{informs2014}
\bibliography{ISR}
\end{document}